Preprint. Under review at IEEE Transactions on Learning Technologies.

# Behaviorally Adaptive Visual Diversion for Inclusive and Resilient Digital Assessment Delivery

Lovi Raj Gupta[1,*], Kamalpreet Kaur[1], Dama Sri Ram[1], Prajithaa Parani[1,1]

[1]*Lovely Professional University, Punjab, India*

**Abstract—Institutions delivering high-stakes digital assessment increasingly rely on layered safeguards — browser lockdown, webcam surveillance, and behavioural analytics — that are typically deployed and evaluated independently of one another and of the learner experience they create. This paper introduces a behaviourally-adaptive visual diversion model for digital assessment rendering, in which a synthetic, non-semantic visual field is composited with assessment content and modulated in real time by observed candidate behaviour. The underlying question content is never altered, reworded, or restructured; only the visual presentation carries an adaptive, low-intensity diversion signal designed to be imperceptible in effect to a genuine candidate while degrading the usability of unauthorized screen capture or sharing. Building on this foundation, we introduce an accessibility-aware attenuation mechanism that reduces or eliminates diversion intensity for candidates with a registered visual-processing accommodation, ensuring the mechanism does not introduce an unintended equity burden. We formalize the approach through a coupled dynamical-systems model — comprising a Diversion Field Generator, a Rendering Tensor, a Behaviour Tensor, a Composite Integrity Functional, and a Multi-dimensional Entropy Functional — and establish supporting theoretical properties for content fidelity, rendering stability, entropy boundedness, and integrity convergence. We discuss the framework's implications for equitable, learner-centred assessment security and its integration into adaptive digital-learning platforms. Two properties of the mechanism are made explicit here that a purely descriptive treatment leaves hidden. The diversion field is keyed to a per-session secret, so its resistance to capture rests on that key rather than on the adversary being unfamiliar with the construction; and attenuation buys accessibility at a measurable cost in capture resistance, which we state as a monotone bound rather than assume away. We refer to the model throughout as BAVD (Behaviourally-Adaptive Visual Diversion).**

## I. INTRODUCTION

Digital assessment platforms now serve a highly heterogeneous learner population, and the integrity mechanisms layered onto these platforms — browser lockdown, webcam proctoring, and behavioural analytics — are rarely evaluated for their effect on that population's experience. Two problems compound one another in current practice. First, these safeguards are typically engineered and validated in isolation, so an institution accumulates several partially overlapping tools rather than one coherent system. Second, because most safeguards are calibrated for a notional average candidate, they can create disproportionate friction or anxiety for learners with disabilities, unstable connectivity, or unconventional testing environments, precisely the

[1] Manuscript submitted for review. This work received no external sponsor or grant funding. (Corresponding author: Lovi Raj Gupta.)

L. R. Gupta, K. Kaur, D. S. Ram, and P. Parani are with Lovely Professional University, Punjab, India (e-mail: loviraj@gmail.com).

L. R. Gupta is the founder of, and K. Kaur, D. S. Ram, and P. Parani are affiliated with, TeachGenie.ai, the organization that develops and commercially operates the MARS assessment platform of which the framework in this article is a component; this is disclosed as a potential competing interest. This article is a theoretical contribution and reports no platform performance data. Generative artificial-intelligence tools were used for language refinement, formatting, and manuscript reorganization, consistent with recognized guidance on responsible generative-AI use in research [20]; all technical content was conceived, verified, and is the responsibility of the authors.

learners most likely to need accommodation.

This paper develops a behaviourally-adaptive visual diversion model that addresses both problems at once. Rather than adding another independent monitoring layer, the model couples a real-time visual-diversion mechanism directly to observed candidate behaviour, so that intensity rises only when behavioural evidence warrants it and falls back to a minimal baseline otherwise. The mechanism operates purely at the level of visual presentation: a synthetic, non-semantic decoy field is composited with genuine content to degrade the usability of unauthorized screen capture or sharing, while the assessment item itself is never reworded, restructured, or otherwise altered. Because the mechanism is intensity-adaptive rather than fixed, it also admits a direct accessibility safeguard: candidates with a registered visual-processing accommodation can have diversion intensity attenuated or removed entirely without weakening the mechanism for the wider cohort.

The remainder of this paper proceeds as follows. Section 2 reviews related work in assessment security and visual-attention theory. Section 3 introduces the system model. Sections 4–9 develop the Diversion Field Generator, the Rendering Model, Behavioural Coupling, the Composite Integrity Functional, the Multi-dimensional Entropy Model, and the state-space formulation. Section 10 introduces the accessibility-aware attenuation mechanism. Section 11 establishes supporting theoretical results, Section 12 discusses implications for equitable deployment, and Section 13 concludes.

## II. RELATED WORK

Assessment-security research has documented the limitations of treating browser lockdown, webcam surveillance, and behavioural analytics as independent safeguards; Dawson [1] argues for coordinated, systemic approaches to assessment integrity rather than an accumulation of point solutions. Keystroke-dynamics research offers a validated, largely unobtrusive behavioural channel for continuous engagement monitoring [2], [3], which this work incorporates as one evidence stream within a broader integrity model. Separately, computational models of visual attention explain why a synthetic visual field can selectively disrupt an unauthorized viewer without disrupting an authenticated one: Itti and Koch's [4] saliency-based account of visual attention shows that conspicuous, high-contrast elements dominate early visual processing, while Treisman and Gelade's [5] feature-integration theory shows that binding several competing visual features into a coherent percept requires serial attentional effort. Together these findings motivate a diversion field whose competing visual elements are costly for an unfamiliar, unsynchronized viewer or automated capture pipeline to parse, while remaining unobtrusive for a candidate whose attention is already anchored on the genuine content.

Prior educational-technology literature on adaptive assessment interfaces has largely focused on adapting item difficulty or pacing to learner performance, rather than adapting the security layer itself to learner behaviour. Likewise, accessibility guidance for digital assessment has focused on interface conformance (contrast, font scaling, screen-reader compatibility) rather than on the behaviour of security mechanisms layered on top of an otherwise accessible interface. This paper connects these threads by treating diversion intensity as a first-class adaptive variable, governed jointly by behavioural evidence and by a candidate's declared accessibility profile.

Two further literatures bear directly on the construction and were absent from the earlier framing of this work. The first is visual secret sharing. Naor and Shamir [6] showed that an image can be split into shares that are individually indistinguishable from noise and that recombine optically rather than computationally. BAVD is not a secret-sharing scheme, since the candidate receives a single composited stream rather than two shares, but it borrows the central idea: the security of the presentation depends on a key held by the legitimate viewer, in the sense argued by Shannon [7], and not on the adversary being unable to guess how the field was built. The second is sampling theory. A screen-capture pipeline samples the display at a fixed rate, whereas the eye integrates it [8], and the gap between those two operations is what the temporal component of the diversion field exploits. Nyquist [9] gives the condition under which a sampled component folds to a visible beat frequency, which we use in Section 4.2.

On the accessibility side, critiques of remote proctoring have argued that surveillance-based integrity tooling distributes its burden unevenly across candidates [10]. That criticism applies to any mechanism that adds sensory load, including this one. It is the reason Section 10 treats attenuation as a first-class part of the model rather than a configuration flag, and the reason Section 10.1 reports what attenuation costs instead of claiming it costs nothing.

## III. SYSTEM MODEL

The assessment session is modelled as a coupled dynamical system with six components: spatial, temporal, diversion, behavioural, rendering, and integrity state. The complete session state is the tuple

$$M = \langle S(t), T(t), D(t), B(t), R(t), I(t) \rangle \quad (1)$$

where D(t), the Diversion State, parameterizes only the statistical properties of a synthetic visual field and never touches assessment content. Figure 1 shows the resulting architecture: spatial and temporal state feed the Rendering Engine directly, the diversion and behavioural domains jointly drive the Diversion Field Generator, and the Integrity Functional closes the loop with feedback into subsequent rendering decisions.

### *A. Threat Model and Trust Assumptions*

The earlier formulation stated a system model but no adversary, which left the fidelity bounds of Equations (4) and (5) without a subject. We state the adversary explicitly. The institution controls the rendering client, either a lockdown browser or an instrumented web client, and a per-session key is delivered to that client at session start over an authenticated channel. Security rests on that key. It does not rest on the adversary being unaware of how the diversion field is constructed, which follows the standard position since Shannon [7].

Three adversaries are in scope. A1 is a screen-capture adversary: software on the candidate machine that samples the framebuffer at rate fs and stores or forwards the frames. A2 is a screen-sharing adversary, which is A1 under a real-time transport constraint, so lossy compression and dropped frames apply as well. A3 is an out-of-band optical adversary, a phone or second camera pointed at the display. A1 and A2 are the adversaries the mechanism is designed against. A3 is only partly addressed, for reasons Section 12 sets out.

Two adversaries are out of scope, and neither is addressable at the rendering layer. One has compromised the client and can read the true content before compositing, at which point no presentation-level defence survives. The other memorises items and reproduces them after the session. Stating these boundaries matters because a reader could otherwise take the fidelity asymmetry of Section 4 as a general confidentiality claim, which it is not.

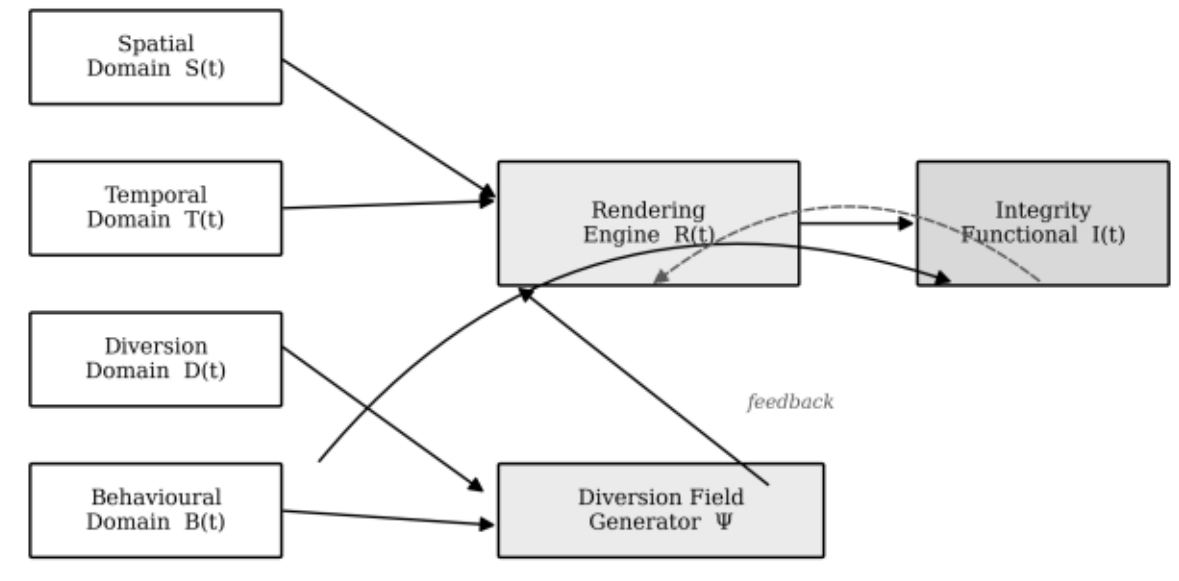

*Fig. 1. Architecture of the behaviourally-adaptive diversion model, showing the diversion domain D(t) and behavioural domain B(t) jointly driving the Diversion Field Generator Ψ.*

## IV. DIVERSION FIELD GENERATOR

The Diversion Field Generator Ψ synthesizes a non-semantic visual field from the diversion state together with the spatial and temporal domains.

Figure 2 traces the pipeline from decoy statistics through spatio-temporal placement to a fidelity-checked composite output.

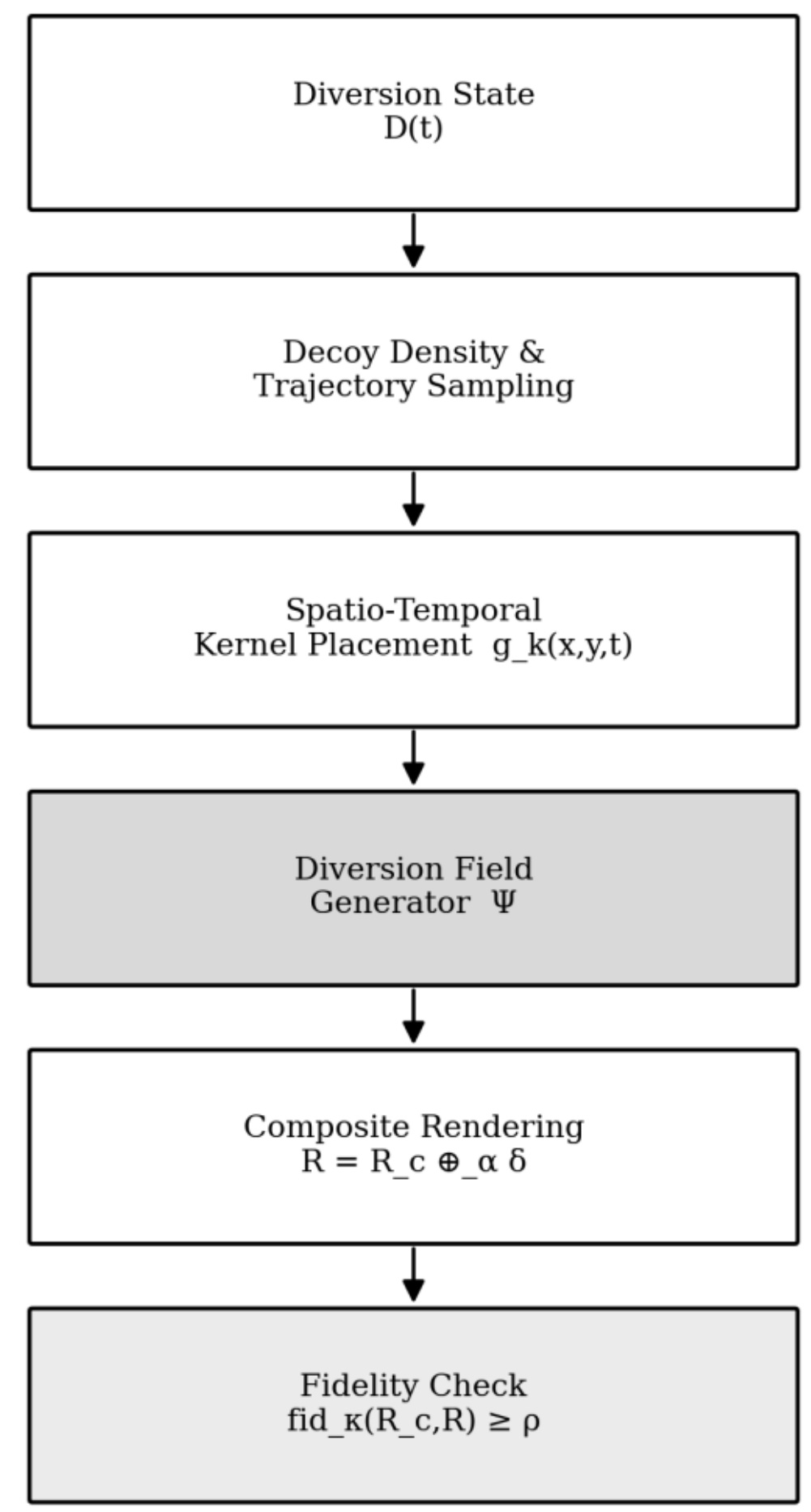

*Fig. 2. Diversion field generation pipeline, from decoy statistics sampled from D(t) to a fidelity-verified composite output.*

The generated field is

$$\delta(x,y,t) = \sum_{k=1}^{N(t)} A_k(t) \cdot g_k(x,y,t) \cdot sin(2\pi f_k t + \varphi_k) \quad (2)$$

summing N(t) decoy elements, each with amplitude Ak(t), a localized spatial kernel gk, and an independent oscillation. The signal presented to the display is the composite

$$R(x,y,t) = R_c(x,y,t) \oplus_\alpha \delta(x,y,t) \quad (3)$$

where Rc is the true, unaltered assessment content, blended with δ at coefficient α(t). Figure 3 shows how this composite is perceived across three viewing conditions: the true content, a synchronized authenticated client, and an unsynchronized capture channel.

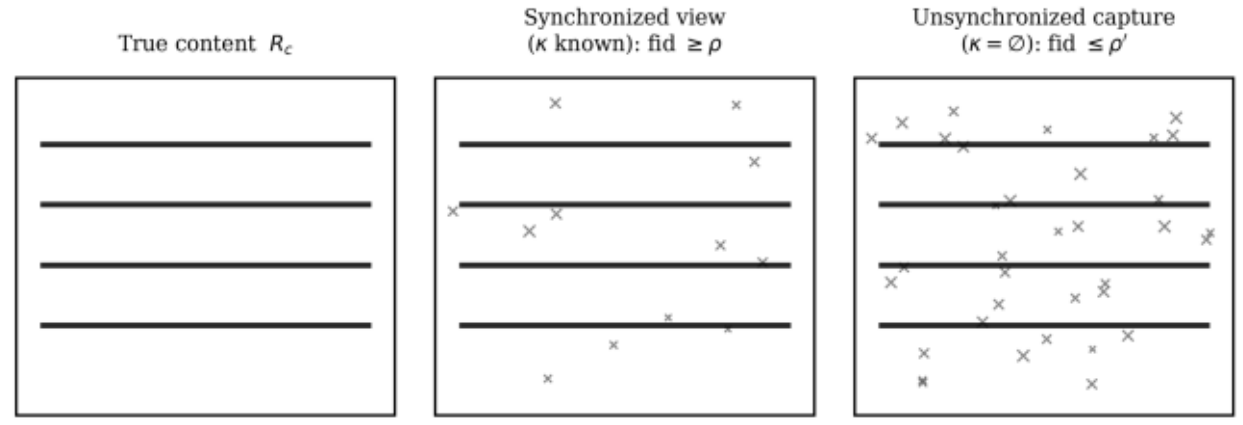

*Fig. 3. Composite rendering as perceived by a synchronized client versus an unsynchronized capture channel, relative to true content Rc.*

A synchronized client, holding the session key κ, recovers content at fidelity

$$fid_\kappa(R_c, R) \geq \rho \quad (4)$$

while any unsynchronized channel is bounded by

$$fid_\emptyset(R_c, R) \leq \rho', \ \rho' < \rho \, (5)$$

For candidates, this means the diversion field is designed to be perceptually negligible during ordinary reading behaviour whenever α(t) remains at or near baseline, which Section 10 ties directly to accessibility accommodation.

*A. Keyed Field Synthesis and the Compositing Operator*

Equation (2) as written is unkeyed. Anyone who learns N(t) and the per-element parameters can regenerate the field and subtract it, so the asymmetry asserted in Equations (4) and (5) does not yet follow from anything. Equation (3) has a second gap: the operator is never defined, so the fidelity functions have no argument to act on. Both are repaired below.

Let F be a pseudorandom function keyed by the session key [11]. Decoy parameters for frame n are derived rather than chosen:

$$\Theta_n = (N_n, \{A_k, x_k, y_k, \sigma_k, f_k, \varphi_k\}) = expand(F_\kappa(sid \,\|\, n)) \quad (2a)$$

where sid is the session identifier and expand maps the function output onto the admissible parameter ranges of Section 5.1. Under the standard PRF assumption, an adversary without the key cannot distinguish the parameter schedule from a uniform draw over those ranges except with negligible advantage. That is the property Theorem 1 needs, and Equation (2) on its own did not supply it.

The compositing operator is alpha blending in a linear light space:

$$R(x, y, t) = (1 - \alpha_{eff}(t))\, R_c(x, y, t) + \alpha_{eff}(t)\, \delta_\kappa(x, y, t) \quad (3a)$$

Written this way the operator is a convex combination, so the composite stays inside the display gamut for every admissible intensity and reduces to the true content exactly when intensity is zero. Content is attenuated in contrast, never overwritten. The per-pixel deviation is then bounded:

$$| R - R_c | \leq \alpha_{eff}(t) \cdot sup\, | \delta_\kappa - R_c | \leq \alpha_{eff}(t) \cdot L_{max} \quad (3b)$$

with L_max the peak display luminance. Equation (3b) is small but load-bearing: it converts Theorem 1 from an assertion into something with a proof, and it is the inequality Section 10.1 differentiates to get the cost of attenuation.

### *B. Perceptual Asymmetry: Integration Against Sampling*

Saliency and feature binding explain why a competing field is expensive for an unfamiliar viewer to parse. They do not explain why the cost should be near zero for the candidate looking at the same screen. Two further mechanisms carry that half of the argument, and both impose design constraints rather than merely describing an effect.

The first is temporal integration. The visual system integrates luminance over a short window and contrast sensitivity falls away steeply at high temporal frequencies [8]. If the field is built so that it sums to zero over any window of length Tv,

$$\int \delta_\kappa(x, y, s)\, ds = 0\ over\ [\, t, t + T_v\, ], for\ all\ x, y, t \quad (4a)$$

then a viewer performing that integration sees the mean of the composite, which by Equation (3a) is the true content. Equation (4a) is a requirement placed on the generator, not a property it has for free. Meeting it forces the decoy set to be partitioned into antiphase pairs with matched duty cycles, and forces the oscillation frequencies above the flicker-fusion range. Section 5.1 shows that this is not always achievable on ordinary hardware.

The second is aliasing. A capture pipeline does not integrate; it samples at rate fs. A component at fk folds to a beat frequency

$$f_{beat} = | f_k - m\, f_s |, m\ chosen\ to\ minimise\ the\ difference \quad (4b)$$

and when fk is placed so that f_beat lands where contrast sensitivity is highest, the captured stream carries the field at close to full amplitude while the live display does not. This is the ordinary moire and refresh-beat effect anyone gets photographing a monitor, used on purpose. Because the parameter schedule of Equation (2a) is keyed and changes every frame, the adversary cannot lock fs to a fixed fk and cancel it, which is precisely what the unkeyed Equation (2) would have permitted.

## V. RENDERING MODEL

Each rendered element, content or decoy, carries an independent rendering state. The Rendering Tensor

$$R_i(t) = \langle x_i, y_i, t_i, \lambda_i, f_i, \varphi_i, \psi \rangle \quad (6)$$

and the associated visibility equation

$$V_i(t) = H\langle sin(2\pi f_i t + \varphi_i + \psi) - \tau \rangle \quad (7)$$

govern when element i appears on screen. Applying distinct frequency and phase settings to decoy elements gives the diversion field motion and flicker characteristics that differ from genuine content, without requiring any change to how content itself is typeset or paced.

### *A. Duty Cycle, Amplitude, and the Safe Frequency Band*

Equation (7) gates element i with a Heaviside threshold on a sinusoid, and its duty cycle follows in closed form from the geometry of the sine:

$$D_i = 1/2 - (1/\pi)\, arcsin(\tau), \tau \in [-1, 1] \quad (7a)$$

A threshold of zero gives a square wave at fifty per cent duty; pushing the threshold toward one makes elements vanish. This matters because the duty cycle and the amplitude together fix the time-averaged luminance the candidate receives, and the zero-mean condition of Equation (4a) is satisfiable only when antiphase pairs share a duty cycle. In practice that pins the threshold at or very near zero.

Frequency is bounded from below by safety rather than by security. WCAG 2.1 Success Criterion 2.3.1 rules out content flashing more than three times per second above the general and red flash thresholds [12], and photosensitive response peaks in the region of 15 to 25 Hz [13]. Combined with the flicker-fusion requirement of Section 4.2, the admissible band is

$$f_k > f_{cff}, and\ f_k \notin [3, 60]\ Hz\ for\ any\ element\ with\ suprathreshold\ contrast \quad (7b)$$

where f_cff is the critical flicker-fusion frequency for the display and viewing conditions in use. There is an unwelcome consequence, and it is worth stating plainly rather than burying. On a 60 Hz panel the highest representable temporal frequency is 30 Hz, which sits below f_cff and inside the photosensitivity band. The temporal channel of Equation (4a) is therefore unavailable on commodity 60 Hz hardware, and on such displays the mechanism has to fall back on static spatial decoys alone, with a correspondingly weaker capture asymmetry. The temporal argument holds at 120 Hz and above. Any deployment claim should say which regime it is in.

Amplitude is bounded from above by the same reasoning. Detection threshold for a target rises with the contrast of a surrounding masker [14], so decoy amplitudes should be held at or below the masking threshold set by the typeset content itself. This gives a principled ceiling on Ak rather than a tuned one, and it is the ceiling under which the perceptual negligibility claim of Section 4 is meant to be read.

## VI. BEHAVIOURAL COUPLING

Candidate behaviour is captured through the Behaviour Tensor

$$(t) = \langle K(t), M(t), T_b(t), F(t), E(t) \rangle \quad (8)$$

comprising normalized keyboard, mouse, tab-transition, focus, and environmental signals. Figure 4 shows how this evidence closes a loop with diversion intensity through the adaptation law

$$\alpha(t) = \alpha_0 + \beta \cdot g(B(t)) \quad (9)$$

so that diversion intensity rises only when behavioural evidence suggests elevated risk, and otherwise relaxes toward a low baseline α0. This adaptivity is what makes the accessibility attenuation of Section 10 meaningful: intensity is already a dynamic, per-candidate quantity rather than a fixed platform-wide setting.

### *A. Dynamics and Stability of the Adaptation Law*

Equation (9) is written as an instantaneous map from behaviour to intensity, but the loop it sits inside is not instantaneous. Intensity changes what is on screen, what is on screen changes how the candidate reads and moves, and that feeds back into the Behaviour Tensor which sets intensity again. A static gain in a loop of this shape can chatter, and chattering intensity is exactly the flashing that Section 5.1 forbids. Replacing Equation (9) with a first-order lag fixes both problems at once:

$$T_\alpha\, d\alpha/dt = -\alpha(t) + [\, \alpha_0 + \beta\, g(B(t)) \,] \quad (9a)$$

The time constant rate-limits intensity, so a jump in the behavioural signal produces a ramp rather than a step, and the flash-rate constraint becomes enforceable at the intensity level and not only at the element level. It also makes the loop analysable. If g is Lipschitz with constant Lg and the behavioural response to on-screen intensity is Lipschitz with constant LB, the closed loop is a contraction whenever

$$\beta\, L_g\, L_B < 1 \quad (9b)$$

in which case a unique equilibrium exists and the loop converges to it [15]. This is a tuning rule with an operational reading: the adaptation coefficient must be kept small relative to how strongly candidate behaviour responds to visual load. It is also the stability property that Theorem 2 was reaching for and did not reach, as Section 11 discusses.

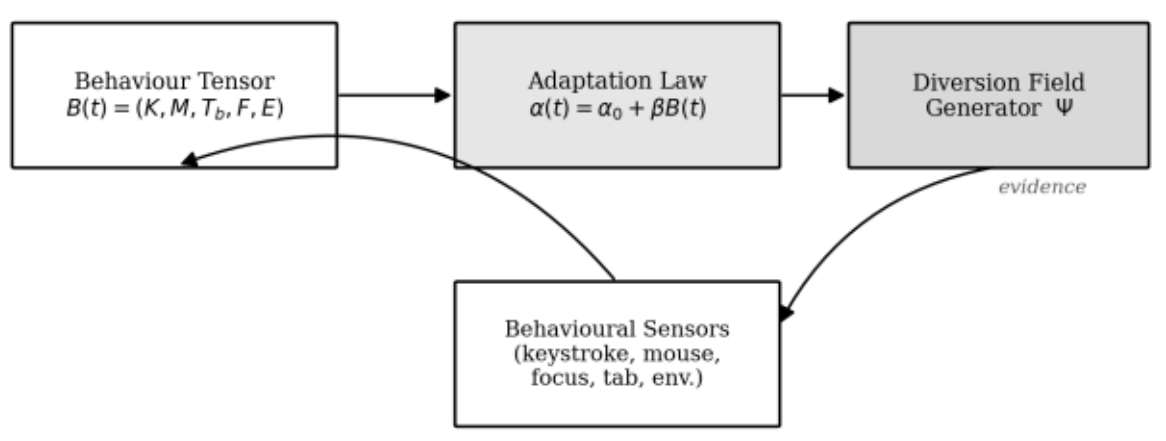

*Fig. 4. Closed-loop coupling between behavioural evidence and adaptive diversion intensity.*

## VII. COMPOSITE INTEGRITY FUNCTIONAL

Session integrity is computed as a weighted fusion of three evidence streams, shown in Figure 5. The Composite Integrity Functional

$$I(t) = w_1 E_b(t) + w_2 E_d(t) + w_3 E_r(t) \quad (10)$$

combines behavioural evidence Eb, diversion-efficacy evidence Ed, and rendering-consistency evidence Er, with weights summing to unity. Institutions can tune w1, w2, w3 to reflect local risk tolerance and pedagogical priorities without altering the underlying detection logic, supporting transparent, auditable integrity decisions rather than an opaque single score.

### *A. Orientation and Dependence in the Evidence Streams*

Two things about Equation (10) need pinning down before it can be used. The first is orientation. The equation does not say which direction the evidence variables run, so the index has no agreed sign. We adopt the convention that each evidence term lies in the unit interval and increases with evidence of an intact session, so an index near one is high confidence and near zero is low. Any implementation must fix this convention before the weights mean anything.

The second is harder. Diversion-efficacy evidence depends on the effective intensity, which by Equation (9a) is driven by the Behaviour Tensor, which is also the source of behavioural evidence. The two streams are therefore correlated, and a weighted sum treats them as though they were not. Weights set as if the streams were independent will overcount whatever the behavioural channel already saw. A conditional form removes the double counting:

$$I(t) = w_1 E_b + w_2 E[ E_d \mid E_b ] + w_3 E[ E_r \mid E_b, E_d ] \quad (10a)$$

at the price of requiring the two conditional expectations to be estimated from session data, which this paper does not have. We flag the problem rather than paper over it. Absent those estimates, Equation (10) should be read as a scoring rule in which the weights encode institutional priority, not as a likelihood in which they encode evidential strength. The distinction changes how an appeal against an integrity decision should be argued, so it is not a technicality.

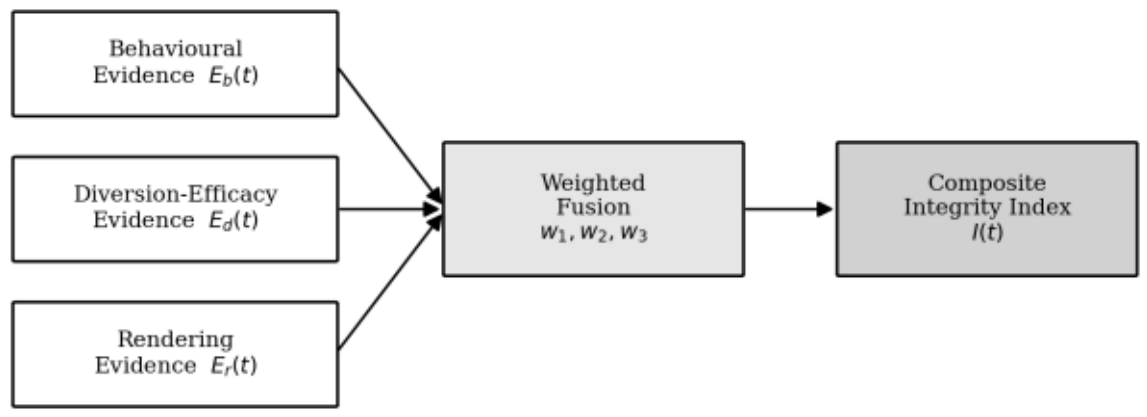

*Fig. 5. Multi-modal evidence fusion for computation of the Composite Integrity Index.*

## VIII. MULTI-DIMENSIONAL ENTROPY MODEL

The unpredictability introduced by the diversion mechanism is quantified with a four-component entropy decomposition,

$$H_{BAVD} = H_S + H_T + H_D + H_B \quad (11)$$

combining spatial, temporal, diversion, and behavioural entropy. This decomposition allows institutions to reason about how much of the system's overall randomness is attributable to the diversion mechanism specifically, as distinct from ordinary behavioural or temporal variation, which is useful when auditing the mechanism for unintended side effects on the candidate experience.

### *A. What the Entropy Decomposition Does and Does Not Measure*

Equation (11) is written as an equality, and as an equality it is wrong unless the four sources are independent. The chain rule gives the correct statement [16]:

$$H(S,T,D,B) = H(S) + H(T|S) + H(D|S,T) + H(B|S,T,D) \leq H_S + H_T + H_D + H_B \quad (11a)$$

so the sum in Equation (11) is an upper bound, and Theorem 3 should be read against Equation (11a) rather than against Equation (11). The gap is not small here, because decoy placement is drawn conditionally on the spatial layout of the item, which makes the diversion and spatial terms dependent by construction. Independence is not a mild idealisation in this system; it is contradicted by the generator.

There is a second and more consequential issue. Marginal entropy is not the quantity a security claim rests on. What an adversary faces is the residual uncertainty about the field after observing the captured stream without holding the key:

$$H_{adv} = H(\, \delta_\kappa \mid R^{cap}, sid \,) \quad (11b)$$

and it is this conditional entropy, not the additive decomposition, that lower-bounds the work of separating true content from the field. Under the assumption of Section 4.1 it is bounded below by the entropy of the parameter schedule, less the adversary distinguishing advantage. The decomposition of Equation (11) keeps a legitimate role: it audits how much on-screen variability the mechanism imposes on a candidate, which is a human-factors quantity. Conflating the two would let a system look secure because it is merely busy. Accordingly the composite entropy is written H_BAVD here; the label MSPDM used in the earlier draft was never expanded and has been retired.

## IX. STATE-SPACE FORMULATION

The assessment process admits an equivalent discrete-time state-space representation. The state equation

$$x_{t+1} = Ax_t + Bu_t + w_t \quad (12)$$

and observation equation

$$y_t = Cx_t + v_t \quad (13)$$

together describe session evolution and the mapping from internal state to observable behaviour [18], [19], shown in Figure 6, enabling future real-time estimation of diversion intensity and integrity trajectory over the course of a session.

### *A. A Concrete Instantiation, and the Limits of the Linear Form*

Equations (12) and (13) leave the transition, input, and observation matrices unspecified, and the adaptation law they are meant to describe contains a nonlinearity in g. The linear model is therefore local at best. For the intensity subsystem, however, the matrices are available in closed form. Discretising Equation (9a) at step Δt with the state taken as intensity gives

$$A = 1 - \Delta t \,/\, T_\alpha, B = \Delta t \,/\, T_\alpha, u_t = \alpha_0 + \beta\, g(B(t)) \quad (12a)$$

which is the scalar case of Equation (12) and is stable for step sizes below twice the time constant. Beyond that the honest position is that the full state is not obviously reconstructible. The integrity index is a scalar readout of three evidence streams, so the observation matrix contributes rank one per timestep and recovering the wider state requires accumulating observations over a window. How well conditioned that accumulation is depends on the fusion weights, and it degenerates as any weight approaches zero. An institution that sets one weight to zero has, without necessarily intending to, made part of the state unobservable.

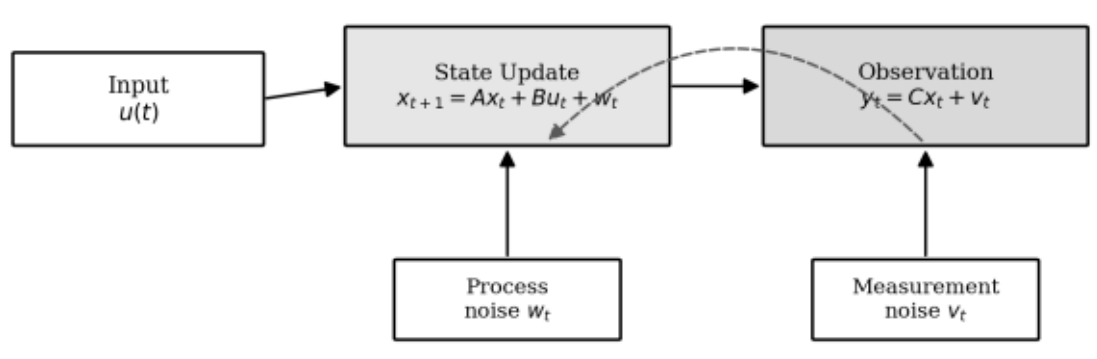

*Fig. 6. Dynamical state-space representation of the assessment-evolution process.*

# X. ACCESSIBILITY-AWARE DIVERSION ATTENUATION

A diversion mechanism that is uniformly intense across all candidates risks becoming an accessibility barrier in its own right, particularly for candidates with visual-processing differences, photosensitivity, or attention-related conditions for whom additional on-screen motion or visual clutter can impose real cognitive or sensory cost. Because diversion intensity in this model is already governed by the adaptive coefficient α(t) rather than fixed at design time, an accessibility accommodation can be incorporated directly into the same adaptation law rather than as a separate opt-out system bolted on afterward.

For a candidate with a registered accommodation profile, we define an accommodation coefficient ai ∈ [0, 1], set through the institution's existing accessibility-declaration process, and apply it multiplicatively to the adapted intensity of Equation (9):

$$\alpha eff\ (t) = \alpha(t) \cdot (1 - a_i) \qquad (14)$$

so that ai = 1 fully suppresses the diversion field for that candidate regardless of behavioural evidence, ai = 0 leaves the standard adaptation law of Equation (9) unchanged, and intermediate values provide graduated attenuation. Figure 7 illustrates the resulting intensity trajectory for a standard candidate against an accommodated candidate under an otherwise identical behavioural evidence stream.

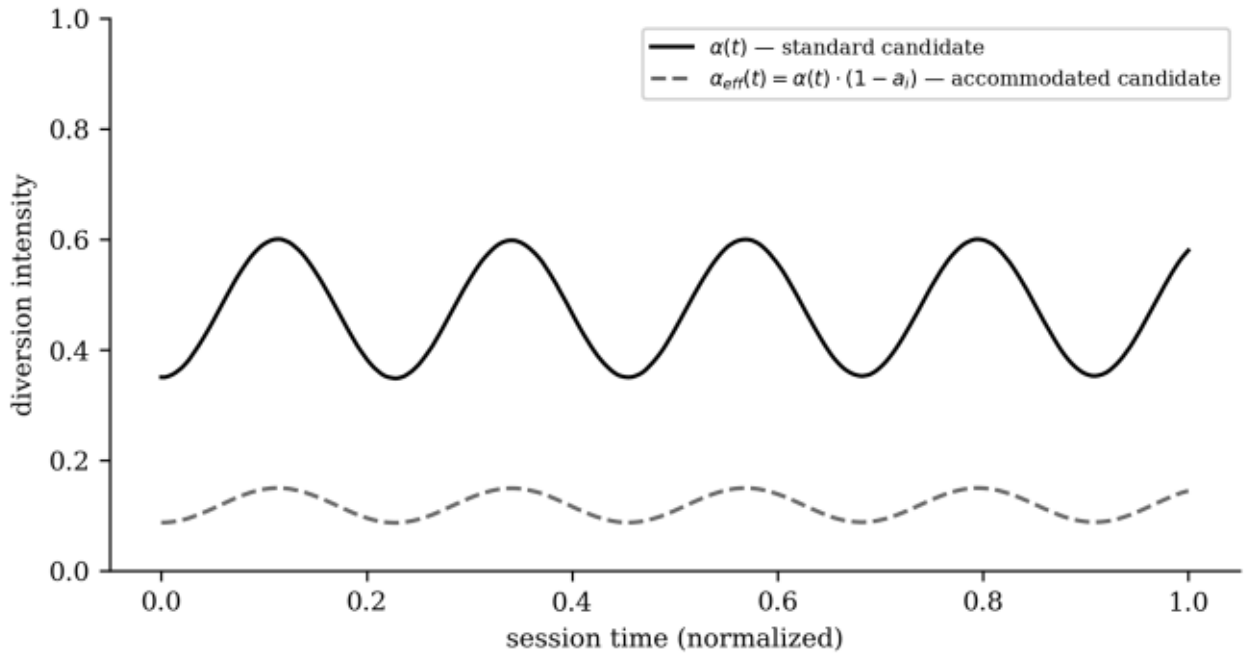

*Fig. 7. Diversion intensity over a session for a standard candidate versus a candidate with a registered accommodation, under the attenuation law of Equation (14).*

Because attenuation acts multiplicatively on α(t) rather than on the underlying behavioural adaptation law, an accommodated candidate's session still contributes evidence to Eb, Ed, and Er in Equation (10); attenuation reduces the visual footprint of the mechanism without exempting the candidate from ordinary integrity evaluation. This separates accessibility accommodation, which concerns the candidate's sensory experience, from integrity evaluation, which concerns evidentiary fusion, avoiding a design in which requesting an accommodation inadvertently weakens the institution's assurance of assessment integrity.

## *A. What Attenuation Costs*

The preceding paragraph is right about evidence collection and incomplete about everything else, and the gap should be closed rather than left. Attenuation preserves content fidelity and rendering stability. It does not preserve capture degradation, which is the property the mechanism exists to deliver. Setting the accommodation coefficient to one drives effective intensity to zero, at which point Equation (3a) returns the true content unmodified and Equation (5) simply fails: an unsynchronised channel now recovers content at the same fidelity a legitimate one does. Attenuation is not free, and describing it as free would misstate the design.

The relationship is at least monotone, which makes it manageable. The deviation available to degrade a capture is proportional to effective intensity by Equation (3b), and capture fidelity decreases in that deviation, so

$$fid_{\emptyset}(R_c, R) \leq \rho'(a_i), with\ \rho'\ nondecreasing\ in\ a_i\ and\ \rho'(1) = \rho \qquad (14a)$$

stated as Proposition 1 in Section 11. The practical reading is that the accommodation coefficient is a residual-risk decision rather than a rendering preference, and an institution granting a coefficient near one owes an account of what replaces the assurance it has given up for that candidate. Reallocating the diversion-efficacy weight of Equation (10) to the other two streams is one answer, since that stream carries almost no information once the field is suppressed. Supervised delivery in a proctored room is another. The model does not choose between them. What it does is make the trade visible and quantified through Equation (14a), instead of leaving it as an unexamined assumption that accessibility and security happen never to conflict.

## XI. THEORETICAL ANALYSIS

The framework is supported by four theoretical properties establishing content fidelity, rendering stability, entropy boundedness, and integrity convergence.

***Theorem 1 (Content Fidelity):*** For a rendering client possessing the session synchronization key κ, $fid_\kappa(R_c, R) \geq \rho$, so an authenticated candidate experiences no material loss of content legibility, and this bound is preserved under accessibility attenuation since $\alpha_{eff}(t) \leq \alpha(t)$ for all $a_i \in [0, 1]$.

***Proof (sketch):***

By Equation (2a) the parameter schedule is a keyed pseudorandom function of the session identifier and frame index, so a client holding the key regenerates the field exactly. By Equation (3a) the composite is a convex combination of true content and field, and the residual after the client accounts for the known field is bounded by Equation (3b) at the effective intensity, which the parameter constraints of Section 5.1 hold below the contrast-masking threshold of the content. The fidelity bound follows from that constraint rather than from construction alone, which is the gap the earlier version of this proof left open. Since accessibility attenuation of Equation (14) can only reduce the effective blend coefficient relative to the unattenuated case, the residual error for an accommodated candidate is no larger than for a standard candidate, so the fidelity bound is preserved, and in fact strengthened, under attenuation. ■

***Theorem 2 (Rendering Stability):*** Composite boundedness and gamut safety. For every admissible effective intensity in the unit interval, the composite of Equation (3a) satisfies $\min(R_c, \delta) \leq R \leq \max(R_c, \delta)$ pointwise, so no pixel is driven outside the display gamut, and the deviation from true content is bounded by Equation (3b). The visibility process of Equation (7) is separately bounded in {0, 1} with duty cycle given by Equation (7a).

***Proof (sketch):***

A convex combination of two quantities lies between them, which gives the pointwise bracket and hence gamut safety; subtracting $R_c$ from Equation (3a) and taking the supremum gives Equation (3b). For the visibility process, the sine is bounded on [−1, 1] for any real argument and the Heaviside operator maps every real input into {0, 1}, so the process is bounded by construction, and the duty cycle follows by measuring the fraction of a period on which the sine exceeds the threshold. The earlier version of this theorem asserted only the {0, 1} bound, which is true by definition of the Heaviside operator and therefore says nothing about the rendering system. The claim above is the one worth making. Note that neither part establishes stability of the adaptation loop; that is Theorem 5. ■

***Theorem 3 (Entropy Boundedness):*** If the four marginal entropies are each finite then their sum is finite, and by Equation (11a) that sum upper-bounds the joint entropy $H(S,T,D,B)$, with equality if and only if the four sources are mutually independent. The composite quantity $H_BAVD$ of Equation (11) is therefore a bound on the system's joint uncertainty rather than a measurement of it.

***Proof (sketch):***

Each component is individually finite by definition, so their sum is finite by closure of addition on the reals. By the sub-additivity of joint Shannon entropy [17], the joint entropy of the four sources is upper-bounded by the sum of their marginal entropies, with equality only under statistical independence, establishing both claims. ■

***Theorem 4 (Integrity Convergence):*** Integrity boundedness and tracking. Let the index be a convex combination of evidence variables in the unit interval with fixed weights summing to one. Then the index is bounded on the unit interval for all t. If each stream is estimated by an exponentially weighted moving average with smoothing parameter λ, and the underlying expectations vary with total variation at most V over a window, then the estimate tracks the time-varying expectation with error bounded by a term of order V/λ plus a stochastic term of order the square root of λ. Boundedness and the tracking bound hold identically for accommodated candidates.

***Proof (sketch):***

Boundedness is immediate: a convex combination of quantities in the unit interval lies in the unit interval. The tracking bound is the standard bias-variance decomposition for an exponentially weighted average, in which the lag term grows with the drift of the underlying expectation and shrinks with the smoothing parameter, while the stochastic term does the reverse; the index inherits both because it is a fixed convex combination of the three estimators. The earlier version of this theorem invoked the law of large numbers to claim almost sure convergence to a fixed limit. That argument requires the evidence streams to be identically distributed over the session, which candidate behaviour is not: it drifts with fatigue and item difficulty, and an adversary has every reason to make it drift on purpose. Under non-stationarity there is no fixed limit to converge to, so the appropriate guarantee is tracking rather than convergence. Attenuation changes only the effective intensity and not the estimator, so the same bound applies to accommodated candidates. ■

***Theorem 5 (Closed-loop Stability of Adaptation):*** Let the adaptation law take the lagged form of Equation (9a) with positive time constant, let g be Lipschitz with constant Lg on the range of the Behaviour Tensor, and let the candidate behavioural response to effective intensity be Lipschitz with constant LB. If the condition of Equation (9b) holds, the closed loop admits a unique equilibrium intensity and converges to it from any admissible initial condition, with the rate of change of intensity bounded by the span of the input divided by the time constant.

***Proof (sketch):***

The composition of the two Lipschitz maps is a contraction on the closed interval of admissible intensities whenever the product of the gain and the two constants is below one, so the Banach fixed-point theorem gives existence and uniqueness of the equilibrium and geometric convergence to it. Convergence of the continuous-time system follows from the lag being a stable first-order filter driven by that contraction [15]. The rate bound is read directly off Equation (9a) by bounding the bracketed input term. That bound is what allows the flash-rate limit of Section 5.1 to be enforced at the intensity level, which the instantaneous form of Equation (9) could not guarantee. ■

***Proposition 1 (Monotone Accessibility-Security Trade-off):*** Let capture fidelity be non-increasing in the magnitude of the deviation between composite and true content. Then under Equation (14a) the attainable capture-degradation threshold is nondecreasing in the accommodation coefficient, and at full accommodation it equals the authenticated fidelity threshold, so the guarantee of Equation (5) is vacuous in that limit.

***Proof (sketch):***

Equation (3b) makes the deviation available for degradation proportional to effective intensity, and Equation (14) makes effective intensity decreasing in the accommodation coefficient. Composing a decreasing map with a non-increasing one gives a nondecreasing capture fidelity bound. At full accommodation the effective intensity is zero, so Equation (3a) returns the true content and both fidelity functions take the same value. The proposition is not a defect in the model; it is the statement that suppressing a visual defence removes the defence, which any deployment granting large accommodation coefficients has to plan around. ■

## XII. DISCUSSION

Coupling diversion intensity to behavioural evidence, and then further to a declared accessibility profile, reframes assessment-security engineering as a design problem with an explicit equity dimension rather than a purely adversarial one. Institutions adopting this model gain a mechanism whose intrusiveness is auditable and tunable per candidate, in contrast to fixed-intensity approaches that impose the same visual load on every test-taker regardless of need. This has practical implications for how institutions communicate assessment-security measures to learners: because the mechanism never alters content and its footprint is demonstrably reduced for accommodated candidates, disclosure to learners can focus on what changes (visual presentation) and what does not (question content, scoring, time allowed).

Several limitations warrant further study. The accommodation coefficient ai is treated here as an institutionally declared input; integrating it with existing accessibility-services workflows, and validating that attenuated diversion intensity meaningfully reduces sensory burden for the intended population, requires dedicated human-subjects study beyond the scope of this formulation. The fidelity function fid(·) is likewise treated abstractly, and its concrete instantiation will shape achievable values of $\rho$ and $\rho'$ in a deployed system. Finally, this paper does not evaluate learner-perceived trust or anxiety associated with an adaptive, behaviourally-responsive security layer, which is an important direction for follow-up work involving learners directly.

Three further limitations deserve stating, because each bounds what the mechanism can be claimed to do. The first is the optical adversary A3 of Section 3.1. A camera photographing the display with an exposure longer than the integration window of Equation (4a) performs the same averaging the eye does, and recovers the content the same way the candidate sees it. The temporal component of the defence is therefore ineffective against a patient photographer, and only the spatial decoys and the attention cost of Section 2 remain. A mechanism that is strong against software capture and weak against a phone camera is worth having, but it should not be sold as more than that.

The second is hardware. As Section 5.1 sets out, the temporal argument requires refresh rates above the flicker-fusion frequency, which rules out the 60 Hz panels most candidates own. Any institution deploying this on bring-your-own-device infrastructure is deploying the weaker spatial-only variant for most of its cohort, and an equity analysis has to account for the fact that display refresh rate correlates with what a candidate can afford.

The third is that no part of this paper has been evaluated empirically. The thresholds, the fidelity functions, the Lipschitz constants of Equation (9b), and the masking ceiling on decoy amplitude are all defined but none are measured. The theoretical results say that the mechanism is well posed and internally consistent under stated assumptions. They do not say that it works, and the distinction should survive into any summary of this work.

## XIII. CONCLUSION

This paper introduced a behaviourally-adaptive visual diversion model for digital assessment delivery in which a synthetic, non-semantic visual field is composited with assessment content and modulated by observed candidate behaviour, without ever altering the underlying question. Building on this coupled dynamical-systems foundation, we introduced an accessibility-aware attenuation mechanism that reduces diversion intensity for candidates with a registered accommodation while preserving the content-fidelity, rendering-stability, and integrity-convergence guarantees established for the general model. Together, these results position visual diversion not merely as a security mechanism but as one that can be engineered for equitable deployment across a heterogeneous learner population.

The framework provides a foundation for future work on empirically validating fidelity and attenuation parameters with real candidates, integrating accommodation declarations with institutional accessibility-services workflows, and extending behaviourally-adaptive security design to other digital-learning contexts beyond summative assessment.

TABLE I
PRINCIPAL SYMBOLS USED IN THE MODEL

| Symbol | Description | Unit / Domain |
|---|---|---|
| M | BAVD state tuple | System tuple |
| S(t), T(t) | Spatial / temporal state | Coord. / time |
| D(t), B(t) | Diversion / behavioural state | Statistical / vector |
| R(t), I(t) | Rendering state / integrity functional | Tensor / scalar |
| Ψ | Diversion Field Generator | Mapping |
| δ(x,y,t) | Generated diversion field | Signal |
| Ak, gk | Decoy amplitude / spatial kernel of element k | Scalar / kernel |
| Rc, R | True content / composite rendered signal | Signal |
| κ | Session synchronization key | Key |
| fidκ(·), fidØ(·) | Fidelity for synchronized / unsynchronized channel | 0–1 |
| ρ, ρ′ | Fidelity threshold / capture-degradation threshold | Scalar |
| Ri(t), Vi(t) | Rendering tensor / visibility of element i | Tensor / binary |
| K, M, Tb, F, E | Keyboard, mouse, tab, focus, env. activity | Normalized |
| α(t), α0, β | Diversion intensity / baseline / adaptation coefficient | Scalar |
| ai, αeff(t) | Accommodation coefficient / attenuated intensity | 0–1 / Scalar |
| Eb, Ed, Er | Behavioural / diversion-efficacy / rendering evidence | Scalar |
| w1, w2, w3 | Evidence weighting coefficients | Scalar |
| HBAVD | Composite model entropy | Bits |
| Θn | Keyed decoy parameter schedule for frame n | Parameter tuple |
| Fκ, sid | Pseudorandom function keyed by κ; session identifier | Mapping / string |
| δκ | Keyed diversion field | Signal |
| Lmax | Peak display luminance | cd/m2 |
| Tv, fcff | Visual integration window; flicker-fusion frequency | s / Hz |
| fs, fbeat | Capture sampling rate; aliased beat frequency | Hz |
| Di, τ | Duty cycle of element i; visibility threshold | 0-1 / [-1,1] |
| Tα | Adaptation time constant | s |
| Lg, LB | Lipschitz constants of g and of behavioural response | Scalar |
| Hadv | Adversary conditional entropy of the field given a capture | Bits |
| HBAVD | Composite model entropy bound | Bits |
| A1, A2, A3 | Capture, sharing, and optical adversaries | Threat class |
| HS, HT, HD, HB | Spatial / temporal / diversion / behaviour entropy | Bits |
| xt, A, But | State vector / transition / input matrix & input | Vector / matrix |
| yt, C | Observation vector / observation matrix | Vector / matrix |

*Note. Rows below the composite-entropy entry were added in this revision, together with the corrected label HBAVD used throughout this table and in Equation (11); the earlier symbol HMSPDM appeared in prior drafts without ever being expanded in the text.*

## Acknowledgment

The authors thank Lovely Professional University for institutional support. No datasets were generated or analyzed in this theoretical study; every parameter reported is a design constraint or a value taken from the cited literature rather than an empirical measurement, and the simulation code used to render the illustrative figures is available from the corresponding author upon reasonable request. A competing-interest disclosure and a statement on the use of generative artificial-intelligence tools in preparing this manuscript appear in the unnumbered footnote on the first page.